\documentclass[sn-mathphys,Numbered]{sn-jnl}
\usepackage{graphicx}%
\usepackage{multirow}%
\usepackage{amsmath,amssymb,amsfonts}%
\usepackage{amsthm}%
\usepackage{mathrsfs}%
\usepackage[title]{appendix}%
\usepackage{xcolor}%
\usepackage{textcomp}%
\usepackage{manyfoot}%
\usepackage{booktabs}%
\usepackage{algorithm}%
\usepackage{algorithmicx}%
\usepackage{algpseudocode}%
\usepackage{listings}%
\usepackage{csvsimple}
\usepackage{pbox}
\usepackage{tabularx, booktabs}
\usepackage{float}
\usepackage{multirow}
\usepackage{makecell}
\usepackage[skip=2pt]{caption}

\begin{document}
\title[Article Title]{Variable Feature Weighted Fuzzy \textit{k}-Means Algorithm for High Dimensional Data}

\author*[1]{\fnm{Vikas} \sur{Singh}}\email{vikkyk@iitk.ac.in}

\author[2]{\fnm{Nishchal K. } \sur{Verma}}\email{nishchal@iitk.ac.in}

\affil[1, 2]{\orgdiv{Electrical Engineering}, \orgname{Indian Institute of Technology}, \orgaddress{ \city{Kanpur}, \postcode{208016}, \state{UP}, \country{India}}}

\abstract{This paper presents a new fuzzy \textit{k}-means algorithm for the clustering of high-dimensional data in various subspaces. Since high-dimensional data, some features might be irrelevant and relevant but may have different significance in the clustering process. For better clustering, it is crucial to incorporate the contribution of these features in the clustering process. To combine these features, in this paper, we have proposed a novel fuzzy \textit{k}-means clustering algorithm by modifying the objective function of the fuzzy \textit{k}-means using two different entropy terms. The first entropy term helps to minimize the within-cluster dispersion and maximize the negative entropy to determine clusters to contribute to the association of data points. The second entropy term helps control the weight of the features because different features have different contributing weights during the clustering to obtain a better partition. The proposed approach performance is presented in various clustering measures (AR, RI and NMI) on multiple datasets and compared with six other state-of-the-art methods.\\

\textit{\textbf{Impact Statement---}} In real-world applications, cluster-dependent feature weights help in partitioning the data set into more meaningful clusters. These features may be relevant, irrelevant, or redundant, but they each have different contributions during the clustering process. In this paper, a cluster-dependent feature weights approach is presented using fuzzy \textit{k}-means to assign higher weights to relevant features and lower weights to irrelevant features during clustering. The method is validated using both supervised and unsupervised performance measures on real-world and synthetic datasets to demonstrate its effectiveness compared to state-of-the-art methods.}

\keywords{\textit{k}-means, Fuzzy \textit{k}-means, cluster validation, sparse data, fuzzy entropy}

\maketitle
\section{Introduction}
\label{intro}

Clustering is a method that tries to organize unlabelled input data points into clusters or groups such that data points within a cluster have the most similarity to those belonging to different clusters, i.e., to maximize the intra-cluster similarity while minimizing the inter-cluster similarity \cite{HCA, bideal}. Based on belongingness, clustering algorithms have been classified into various categories: hierarchical, density-based,grid-based, partition-based, spectral and model-based clustering \cite{bezdek, hierarchical, density, spectral, grid}. Among them, partition-based clustering is the most widely studied in the literature. The most popular partition-based \textit{k}-means algorithms reported in the literature are well-known for their performance in clustering large-scale datasets. However, these algorithms are susceptible to the initial cluster centers \cite{Ac,Ad,ball}. Ruspini \cite{ruspini} and Bezdek \cite{bezdek1980} have presented fuzzy variants of the k-means algorithm, where each data point can be a subset of multiple clusters with a different membership degree.  These membership degrees play a crucial role in addressing the uncertain behavior of both numerical, categorical, image datasets, among others \cite{viknano, ieetfs}.  
The fuzzy \textit{c}-means (FCM) \cite{fcm} and improved FCM \cite{Imfcm} clustering algorithms have been explored more due to their simplicity and data handling characteristics. However, the major problem with the \textit{k}-means (KM) and fuzzy \textit{k}-means algorithm remains the same, i.e., all features are assumed equally crucial during the clustering. As a result, these algorithms are easily affected by outliers.

The weighted feature techniques were studied to overcome the above limitations by assigning different feature weights based on the features’ usefulness in identifying the clusters. The feature-weighted clustering algorithms, i.e., weighted \textit{k}-means (WKM) \cite{wkm}, entropy-weighted \textit{k}-means (EWKM)  \cite{ewkm}, sparse \textit{k}-means (SKM) \cite{fs}, weighted fuzzy \textit{c}-means (WFCM) \cite{wfcm}, feature-weighted FCM based on simultaneous clustering and attribute discrimination (SCAD) \cite{scad}, and feature reduction-based FCM \cite{frfcm}, have been studied. The WKM is an extension of KM in which features are assigned equal weights across all clusters. EWKM is the entropy-weighted clustering method in which features are assigned distinct weights in the various clusters. SKM uses the $L_1$ norm in the objective function, making the feature weights zero when the feature weights are smaller than a predefined threshold. To improve the learning performance of FCM, WFCM is proposed,  in which the feature weights are learned through the gradient descent method. SCAD is presented to improve  FCM performance by simultaneously clustering the data with attribute discrimination.

Variable weighting of features is an important research area in the machine learning domain, which automatically determines the weight of each feature based on its significance \cite{ singh, fwk, nor}. Annette and Frank \mbox {\cite{wfc}} have presented an objective function-based fuzzy clustering technique that helps to find the effect of every single data variable on each cluster. In \mbox {\cite{fsc}}, the author has presented a novel method for feature selection during the clustering based on an idea to transfer an alternative to the fuzzifier, which controls the membership degrees’ influence to attribute weights. In \mbox {\cite{fcer}}, a new exponential form of the fuzzy memberships was introduced to ensure the consistency bounds and makes it possible to interpret the mixing parameter as the variance of the cluster. In  \mbox {\cite{pfsc}}, the authors have solved the subspace clustering problem by minimizing the FCM cost function with a weighted Euclidean distance and a non-differentiable penalty term. For learning the number of clusters (\textit{k}), the relevance of individual features in each cluster has been considered. The problem of \textit{k}-means and fuzzy \textit{k}-means algorithms, where all features are considered equally important during the clustering process, is minimized. Specifically, for handling the sparse and high dimensional data, each feature’s contribution is essential; although some features may be irrelevant, they still hold different significance in the clustering process \cite{tw, agglomerative}.

This paper presents a novel entropy-based, variable feature weighted fuzzy \textit{k}-means clustering by considering each cluster with different feature weights for an optimal partition. Since each feature has different contributions to identify the data points in a cluster during the clustering process, the difference in these features’ contribution is formulated in terms of weight that can be expressed as the degree of certainty in the cluster.  In the clustering process, a decrease in weighted entropy within a cluster indicates an increase in the certainty of a subset of features that have more substantial weights in determining the cluster.  However, membership entropy helps identify the data points in a cluster and makes clustering insensitive to the initial cluster centers. In the proposed approach, we have simultaneously minimized inter-cluster dispersion, maximizes the negative data points-to-clusters membership entropy, and negative weight entropy to enhance the number of features identifying the number of clusters.

The major contributions are briefly summarized as follows:
\begin{enumerate} 
\item For handling the high dimensional, sparse, and noisy data, the fuzzy \textit{k}-means objective function is modified using the fuzzy membership entropy with weight entropy to increase the robustness of the clustering algorithm.
\item The fuzzy membership entropy  helps in minimizing the within-cluster dispersion and maximize inter-cluster dispersion to identify the data points in a cluster.
\item The weighted feature entropy helps to simulate more dimensions to add different contributions to  identify the number of optimal clusters.
\end{enumerate}

The rest of the paper is organized as follows: The proposed methodology is briefly described in  Section \ref{problem formulation}. Experimentation and comparisons of the proposed approach with various state-of-the-art methods on both real-world and synthetic datasets are discussed in Section \ref{results and discussion}. Finally, Section \ref{conclusion} concludes the complete paper.

\section{Proposed Methodology}
\label{problem formulation}
This section presents a novel entropy-based variable feature weighted fuzzy \textit{k}-means clustering algorithm, considering each feature's different contributions in each cluster. In the proposed approach, the weighted fuzzy \textit{k}-means algorithm is combined with fuzzy membership entropy and weighted  feature entropy.  {As defined in  \mbox {(\ref{mainob})}, the first term measures the dissimilarity between the samples within clusters, and the second term measures the membership entropy between samples and clusters during the clustering process. As shown in \mbox {(\ref{mainob})}  if $\mu_{ij}$ is close to zero, then the second term will be close to zero, which means the $i^{th}$ data point is not assigned to $j^{th}$ cluster . However, if $\mu_{ij}$  close to one, $i^{th}$ data point is  assigned to $j^{th}$ cluster with value depending on $\lambda_{i}$. Finally, the third term, i.e., the feature weights' entropy, represents the degree of certainty to the features in identifying the cluster. Similarly,  $\mu_{ij}$,  $w_{jl}$ is close to zero, then the third term will be close to zero, which means the $l^{th}$ feature is not assigned to $j^{th}$ cluster. However, if $w_{jl}$  close to one $l^{th}$ feature is  assigned to $j^{th}$ cluster with value depending on $\gamma_{j}$. }

Let us consider a data matrix   $X = [X_1, X_2,\dots, X_n] \in R^{m\times n}$, where $m$ and $n$ are the number of features and number of samples, respectively. Here, $X_i = [x_{i 1}, x_{i 2}, \dots, x_{i m}]' \in R^{m}$ represents  the $i^{th}$ sample in the data matrix. To group the data matrix $X$ into $k$ number of clusters, the  following objective function can be minimized
\begin{align}
\label{mainob}
\begin{split}
    P(U, V, W) = \sum_{i=1}^n\sum_{j=1}^k\sum_{l=1}^m \mu_{i j} w_{j l}D_{i j}^{l} + \sum_{i=1}^n \lambda_{i} \sum_{j=1}^k \mu_{i j} \log \mu_{i j} \\ + \sum_{j=1}^k \gamma_{j} \sum_{l=1}^m w_{j l}\log w_{j l}
\end{split}
\end{align}
subject to 
		\begin{align}
\begin{split}
    \sum_{j=1}^k\mu_{i j}= 1,\,\,  0\leq \mu_{i j} \leq 1, \,\, 1\leq i \leq n \\
    and \\
    \sum_{l=1}^m w_{j l}= 1,\,\,  0\leq w_{j l} \leq 1, \,\, 1\leq j \leq k
\end{split}
\end{align}

where, $D_{i j}^{l} = (x_{i l}-v_{j l})^2,$ is the dissimilarity measure between $i^{th}$ sample and $j^{th}$ cluster, $x_{i l}$   is the value of $l^{th}$ feature of $i^{th}$ sample, $v_{j l}$ is the value of $l^{th}$ feature of $j^{th}$ cluster, $U=[u_{i j}]$ is the $n\times k$, fuzzy partition matrix, where $u_{i j}$ is  the membership degree value of the $j^{th}$ cluster of $i^{th}$ sample, $V = [v_{j l}]$  is the $k\times m$, matrix having	 the cluster centers, and $W = [w_{j l}]$ is an $k\times m$, weight matrix, $w_{j l}$ is the weight value of  $l^{th}$ feature to $j^{th}$ cluster, $\gamma_{i}$ and $\lambda_{j}$ are input parameters used to control the fuzzy partition and feature weight, respectively.

  \subsection{Optimization Procedure}
      The objective function, as given in (\ref{mainob}), is a constrained nonlinear optimization  whose
  solutions are unknown. The main aim is to minimize $P$ with respect to $U$, $V$,  and $W$ using alternating optimization method. In the  alternating optimization method,  first we fix $U=\hat{U}$ and $W = \hat{W}$  and minimize $P$ with respect to $V$. Then we fix $U=\hat{U}$ and $V=\hat{V}$  and minimize $P$ with respect to $W$. Afterward, we fix $V=\hat{V}$ and $W=\hat{W}$  and minimize $P$ with respect to $U$.

  First for the fixed $U$ and $W$  the objective function $P(U, V, W)$ is minimized with respect to $V$ as
  \begin{align}
  \label{center}
      \frac{\partial \hat{P}(U, V, W)}{\partial \hat{v}_{j l}}  = -2\sum_{i=1}^n \mu_{i j} w_{j l} (x_{i l}-v_{j l})= 0
  \end{align}

  From (\ref{center}) the cluster center can be obtained using
  \begin{align}
      \label{ficenter}
      v_{j l}  = \frac{w_{jl}\sum_{i=1}^n\mu_{i j} x_{i l}}{w_{jl}\sum_{i=1}^n \mu_{i j}}
  \end{align}
  From the above equation two cases are arises.

  Case 1: If $w_{jl}=0 $, the $l^{th}$ feature is totally irrelevant respective to the $j^{th}$ cluster. Hence, for any value of $v_{j l}$,  this feature will not contribute to determining  the weighted distance. Therefore, in this case, any random value of $v_{j l}$ can be selected. 

  Case 2: If $w_{jl}\neq 0 $, the $l^{th}$ feature is relevant respective to the $j^{th}$ cluster, then the (\ref{ficenter}) is written as

  \begin{align}
      \label{fincenter}
      v_{j l}  = \frac{\sum_{i=1}^n\mu_{i j} x_{i l}}{\sum_{i=1}^n \mu_{i j}}
  \end{align}

  Then for given $V = \hat{V}$, the constraint optimization problem in (\ref{mainob}) is changed into unconstrained minimization problem using Lagrangian multiplier technique as follows:
  \begin{align}
  \label{parti}
  \begin{split}
      \hat{P}(U, W, \alpha, \delta) = \sum_{i=1}^n\sum_{j=1}^k\sum_{l=1}^m \mu_{i j} w_{j l}D_{i j}^{l} +  \sum_{i=1}^n \lambda_{i}\sum_{j=1}^k \mu_{i j} \log \mu_{i j}\\ +  \sum_{j=1}^k \gamma_{j}\sum_{l=1}^m w_{j l}\log w_{j l}-\sum_{i=1}^n\alpha_i\left(\sum_{j=1}^k \mu_{i j} -1\right) \\ - \sum_{j=1}^k\delta_j\left(\sum_{l=1}^m w_{j l} -1\right) 
  \end{split}
  \end{align}
  where,  $\alpha = [\alpha_1, \alpha_2,\dots, \alpha_n]$ and $\delta = [\delta_1, \delta_2,\dots, \delta_k]$ are the vectors containing the Lagrangian multipliers.  If $(\hat{U}, \hat{W}, \hat{\alpha}, \hat{\delta})$ are the optimal values of $\hat{P}(U, W, \alpha, \delta)$, then the gradient with respect to these variable are vanishes  and they are  written as 
  \begin{align} 
  \label{diff_mu}
  \begin{split}
      \frac{\partial \hat{P}(U, W, \alpha, \delta)}{\partial \hat{\mu}_{i j}} = \sum_{l=1}^m w_{j l} D_{i j}^{l} +\lambda_{i} (1+\log\mu_{i j}) -\alpha_{i} =0\\
      1\leq i \leq n,\,\,  1\leq j \leq k,
      \end{split}
  \end{align}

  \begin{align} 
  \label{diff_w}
  \begin{split}
      \frac{\partial \hat{P}(U, W, \alpha, \delta)}{\partial \hat{w}_{j l}} = \sum_{i=1}^n \mu_{i j} D_{i j}^{l} +\gamma_{j}(1+ \log w_{j l}) -\delta_{j} =0 \\
      1\leq j \leq k,\,\,  1\leq l \leq m,
      \end{split}
  \end{align}

  \begin{align} 
  \label{diff_al}
      \frac{\partial \hat{P}(U, W, \alpha, \delta)}{\partial \hat{\alpha}_{i}} = \sum_{j=1}^k \mu_{i j} -1 =0  
  \end{align}
  and,
  \begin{align}
  \label{diff_del}
      \frac{\partial \hat{P}(U, W, \alpha, \delta)}{\partial \hat{\delta}_{j}} = \sum_{l=1}^m w_{j l} -1 =0  
  \end{align}
   the (\ref{diff_mu}) and (\ref{diff_w}) can be simplified as
   \begin{align}
   \label{mu}
       \mu_{i j} = \exp \left(\frac{-\sum_{l=1}^m w_{j l} D_{i j}^{l}}{\lambda_{i}}\right)\exp\left( \frac{\alpha_{i}}{\lambda_{i}}\right)\exp(-1)\\
   \label{wl}
       w_{j l} = \exp \left(\frac{-\sum_{i=1}^n \mu_{i j} D_{i j}^{l}}{\gamma_{j}}\right)\exp\left( \frac{\delta_{j}}{\gamma_{j}}\right)\exp(-1)
   \end{align}

  By substituting (\ref{mu}) in (\ref{diff_al}) and (\ref{wl}) in (\ref{diff_del}), we have 
  \begin{align}
  \label{simmu}
  \begin{split}
       \sum_{j=1}^k \mu_{i j} =\sum_{j=1}^k \exp \left(\frac{-\sum_{l=1}^m w_{j l} D_{i j}^{l}}{\lambda_{i}}\right)\exp\left( \frac{\alpha_{i}}{\lambda_{i}}\right)\exp(-1)\\ 
     =\exp\left( \frac{\alpha_{i}}{\lambda_{i}}\right)\exp(-1)\sum_{j=1}^k \exp \left(\frac{-\sum_{l=1}^m w_{j l} D_{i j}^{l}}{\lambda_{i}}\right)=1 \\
       \exp\left( \frac{\alpha_{i}}{\lambda_{i}}\right)\exp(-1)=\frac{1}{\sum_{j=1}^k \exp \left(\frac{-\sum_{l=1}^m w_{j l} D_{i j}^{l}}{\lambda_{i}}\right)}
  \end{split}
  \end{align}
  Similarly,
  \begin{align}
  \label{simlw}
      \begin{split}
          \sum_{l=1}^m w_{j l} =   \sum_{l=1}^m \exp \left(\frac{-\sum_{i=1}^n \mu_{i j} D_{i j}^{l}}{\gamma_{j}}\right)\exp\left( \frac{\delta_{j}}{\gamma_{j}}\right)\exp(-1) \\
          = \exp\left( \frac{\delta_{j}}{\gamma_{j}}\right)\exp(-1)\sum_{l=1}^m \exp \left(\frac{-\sum_{i=1}^n \mu_{i j} D_{i j}^{l}}{\gamma_{j}}\right) =1\\
              \exp\left( \frac{\delta_{j}}{\gamma_{j}}\right) \exp(-1)=\frac{1}{\sum_{l=1}^m \exp \left(\frac{-\sum_{i=1}^n \mu_{i j} D_{i j}^{l}}{\gamma_{j}}\right)}
      \end{split}
  \end{align}

  By substituting (\ref{simlw}) in (\ref{wl}) and  (\ref{simmu}) in  (\ref{mu}), we obtain
  \begin{align}
      \label{fiwl}
      w_{j l} = \frac{\exp \left(\frac{-\sum_{i=1}^n \mu_{i j} D_{i j}^{l}}{\gamma_{j}}\right)}{\sum_{l=1}^m \exp \left(\frac{-\sum_{i=1}^n \mu_{i j} D_{i j}^{l}}{\gamma_{j}}\right)}\\
      \label{fimu}
      \mu_{i j} = \frac{ \exp \left(\frac{-\sum_{l=1}^m w_{j l} D_{i j}^{l}}{\lambda_{i}}\right)}{\sum_{j=1}^k \exp \left(\frac{-\sum_{l=1}^m w_{j l} D_{i j}^{l}}{\lambda_{i}}\right)}
  \end{align}
As shown in  (\ref{fiwl}) and (\ref{fimu}) the feature weights within the cluster and membership degree of the data points are depend on each other. The dependency of both term on each other helps to find the better partition of the dataset and makes the clustering algorithm  insensitive to the initial cluster centers. 
\subsection{Parameter Selection}
The choice of parameters $ \gamma_{j}$ and $ \lambda_{i}$ in (\ref{fiwl}) and  (\ref{fimu})  is essential to the performance since they reflect the importance of the second and third term relative to the first term in (\ref{mainob}). The parameter $\lambda_i$ controls the clustering process in two ways: First, when $\lambda_i$  is large such that the first term in (\ref{mainob}), i.e., within-cluster dispersion, is small in comparison to the second term, the second then plays a crucial role in minimizing (\ref{mainob}). In the clustering process, it tries to assign data points to more than one cluster to make the second term more negative. When the membership degrees $\mu_{i j}$ of a data point to all clusters are the same, the membership entropy value is large. Since the locations of points are fixed, to get the largest entropy, the cluster centers should be moved to the same location. Second, when $\lambda_i$ is small, the first term is large, and it will play the role of minimizing the within-cluster dispersion. However, the control parameter $\gamma_j$ is used to control the feature weights, when $\gamma_j$ is positive, then the weight $w_{j l}$ as given in (\ref{fiwl}) is inversely proportional to $\sum_{i=1}^n \mu_{i j} D_{i j}^{l}$.  The small of this term will make large $w_{j l}$, i.e., $l^{th}$ features in $j^{th}$ cluster are more important. If $ \gamma_{j}$ in (\ref{fiwl}) is too large,  the third term will dominate, and all features in $j^{th}$  cluster will be relevant and assigned equal weights of $\frac{1}{m}$.  The parameters  $ \lambda_{i}$ and $ \gamma_{j}$ in (\ref{fiwl}) and (\ref{fimu}) are computed by  (\ref{lambda}) and  (\ref{gamma}) in iteration, \textit{t}, as
    \begin{align}
    \label{lambda}
    \lambda_{i} =K_{1} \frac{\sum_{j=1}^k\sum_{l=1}^m \mu_{i j}^{t-1} w_{j l}^{t-1}(D_{i j}^{l})^{t-1}}{\sum_{j=1}^k  -\mu_{i j}^{t-1} \log \mu_{i j}^{t-1} +\eta}
    \end{align}
    \begin{align}
    \label{gamma}
    \gamma_{j} = K_{2}\frac{\sum_{i=1}^n\sum_{l=1}^m \mu_{i j}^{t-1} w_{j l}^{t-1}(D_{i j}^{l})^{t-1}}{\sum_{l=1}^m w_{j l}^{t-1} \log w_{j l}^{t-1} +\eta}
    \end{align}
{The parameter $K_1$ and $K_2$ are a constants, and the superscript ($t-1$)  represents the values in iteration ($t-1$). The parameter $\eta$ is chosen to be very small values when the data points belong to single cluster or only single feature is important in the dataset}. The presented approach is briefly described in Algorithm \ref{FTD-means} as follows:
\begin{algorithm}
\caption{An Entropy based Variable Weighted Fuzzy \textit{k}-Means  Clustering Algorithm}
\label{FTD-means}
\begin{algorithmic}[1]
\item Give the number of cluster $k,$ randomly initialize cluster
center $V^{(0)}$ using maxi-min method i.e., $\min(\min(X)) +(\max(\max(X))-\min(\min(X)))*rand(n,k)$,   initialize feature weight $W^{(0)}$ as $\frac{1}{m}*ones(k, m)$  and initialize randomly control parameter $\lambda^0$ and $\gamma^0$
\State\textbf{Repeat}
  \item Update the feature weight matrix $W$ using (\ref{fiwl})
 \item Update the fuzzy partition matrix $U$ using (\ref{fimu})
  \item Update the cluster center matrix $V$ using (\ref{fincenter})
  \item Update parameters  $\lambda_{i}$ and $\gamma_{j}$ using (\ref{lambda}) and (\ref{gamma})
     
\State\textbf{Until} The objective function achieves local minima or centers stabilize
 \end{algorithmic}
\end{algorithm}
\begin{table}[ht]
\centering %
\caption{\textsc{  \small dataset descriptions }} %
\begin{tabular}{ |c|c|c|c|}
\hline
  Datasets  & \# of samples & \#  of features & \#  of clusters  \\
\hline
IRIS \cite{Dua}  &  $150$  & $4$ & 3  \\ 
\hline
Ionosphere \cite{Dua} &  $351$  & $34$ & 2 \\ 
\hline
Ovarian cancer \cite{ovar}&  $253$  & $15154$ & 2  \\ 
\hline
Colon cancer \cite{colon} &  $62$  & $2000$ & 2\\ 
\hline
Ovarian cancer \cite{ova} &  $216$  & $ 4000$ & 2 \\ 
\hline
Glioma  \cite{data} & $50$  & $4434$ & 4  \\ 
\hline
 Lung  cancer \cite{data}&  $203$  & $3312$ & $5$ \\ 
 \hline
  Wine \cite{Dua}&  $178$  & $13$ & $3$ \\ 
 \hline
\end{tabular}
\label{Table: dataset}
\end{table}
\begin{table*}[ht]
\centering %
\caption{\textsc{  \small performance measure in terms of AR, RI, and NMI (in \%)}} %
\resizebox{0.68\textheight}{!}{%
\begin{tabular}{|c|c|c|c|c|c|c|c|c|}
\hline
 Datasets &   & {KM \cite{km}}  &{EWKM \cite{ewkm}}  & {AFKM \cite{agglomerative}} &{FCM \cite{fcm}} & {SCAD \cite{scad}} & {GMM \mbox {\cite{gmm}}}&{Proposed Approach}\\
\hline
 \multirow{3}{*}{Iris \cite{Dua}} &            AR   &    88.67         &  89.78  &  90.67          &         82.67      &88.67  &  96.00 & \textbf{96.00}   \\

&           RI      &    87.37       &88.39  &   88.93       &  82.78 & 87.37  &  94.23  &   \textbf{94.23} \\

&        NMI         &  74.19        &  77.80   &     72.44       & 65.22 &  74.19 &  86.96 & \textbf{88.91}\\

\hline 
 \multirow{3}{*}{Ionosphere \cite{Dua}} &         AR    &    64.10           &  71.08  &      70.37   &      54.42      &  70.94 & 58.97   & \textbf{71.23}  \\

&           RI    &  53.85         & 58.77   &     58.18       & 50.25  &  58.65 &  51.47   & \textbf{58.89}\\

&        NMI       &  00.00       &   13.35   &     12.64       &  02.53 &  12.99 &  00.66  & \textbf{13.49} \\

\hline
 \multirow{3}{*}{Ovarian cancer \cite{ovar}} &            AR     &     55.73         &   55.73   &   55.73   &         64.81   &  56.13 &  64.10 &  \textbf{81.03}   \\

&           RI   & 50.46           & 50.46 &         50.46   & 54.18 &50.56 &  53.84    &  \textbf{69.13}  \\

&        NMI           & 00.71       &     00.71  &         07.00   & 07.44 &  00.95&  06.79   & \textbf{27.11}\\

\hline
 \multirow{3}{*}{Colon cancer \cite{colon}} &        AR  &       51.61         &   53.23  &   53.23    &       51.61     &  53.23 & 53.12      &  \textbf{55.65} \\

&           RI    &   49.23           &  49.39 &      49.39      & 49.23  &  49.39 &  47.44   & \textbf{49.84}   \\

&        NMI      & 02.27        &     04.59   &     01.27       & 01.27  &   04.59&    4.32   & \textbf{05.33}  \\

\hline
 \multirow{3}{*}{Ovarian cancer \cite{ova}} &        AR      &       56.02        &   76.85   &  71.29    &        64.81    &  71.30 &    76.34 & \textbf{86.29}  \\

&           RI   &   50.50         &  64.25  &         58.88   & 54.18 & 58.88 &    72.10  & \textbf{82.83}   \\

&        NMI        &  00.00        &  32.24   &       18.19     & 07.44 &  18.76 &  43.33   & \textbf{46.95} \\



\hline
 \multirow{3}{*}{Glioma  \cite{data}} &          AR          &  48.00          & 64.00 &  62.00          &     56.00       &  54.00  &   55.45   & \textbf{66.00}\\

&           RI&      70.12         &       74.37    &      73.96      &70.94  &72.65 &  72.93  & \textbf{75.35}  \\

&        NMI  &    44.45           &      \textbf{51.86}      &   49.41     &46.74 &  49.61  &  46.34 &47.86  \\

\hline
 \multirow{3}{*}{Lung cancer \cite{data}} &         AR    &  49.75         & 62.44  &     51.72   &       55.56        &59.61& 61.56 &  \textbf{64.04}      \\

&           RI    &     61.09       &\textbf{64.11} &58.98         &    51.48 & 56.67 &  57.32  & 58.57    \\

&        NMI      &      45.75      & \textbf{49.16}& 42.50           & 27.33 & 19.99 &   21.39  & 26.86 \\



\hline


 \multirow{3}{*}{Wine \cite{Dua}} &         AR    &    57.30           &  59.66     & 69.66     &  69.66                 &69.66    &  69.66    &  \textbf{70.22}      \\

&           RI    &    69.18      &71.35&  71.35  &  71.35       & 71.35 & 69.25   & 71.87   \\

&        NMI      &    42.41        & 42.12  &  42.12     &  42.44           &  42.12  &   41.48  & 42.88 \\
 \hline
\end{tabular}  
}
\label{Table: 2}
\end{table*}
\begin{table*}[ht]
\centering %
\caption{\textsc{  \small performance measure in terms of PC, CE, XB and DI }} %
\resizebox{0.68\textheight}{!}{%
\begin{tabular}{|c|c|c|c|c|c|c|}
\hline
 \multirow{2}{*}{Datasets} & &     & {AFKM \cite{agglomerative}} &{FCM \cite{fcm}}  & SCAD\cite{scad} &{Proposed Approach}\\

 &  \makecell{ True \\ cluster}  & \makecell{ Validity \\ indices} &  \makecell{Optimal cluster\\(values)} &  \makecell{Optimal cluster\\(values)}  &    \makecell{Optimal cluster\\(values)} & \makecell{Optimal cluster\\(values)}\\
\hline
 \multirow{4}{*}{Iris \cite{Dua}} &  \multirow{4}{*}{3} &         PC  &    3 (0.798)      &  3 (0.783) & 3 (0.783) &    \textbf{3 (0.822)} \\

&  &         CE &  3 (0.299)  &  3 (0.395) & 3 (0.395)&    \textbf{3 (0.294)} \\

 &  &        XB  &  3 (3.614) & 3 (4.231) & 3 (4.230)  &   \textbf{3 (3.575) }\\

& &         DI  & 3 (0.030) &  \textbf{3 (0.105)} &  \textbf{3 (0.105)} &   3 (0.052)\\
\hline
 \multirow{4}{*}{Ionosphere \cite{Dua}} & \multirow{4}{*}{2} &         PC  & 2 (0.500)   &  2 (0.651) &  2 (0.651) &    \textbf{2 (0.729)} \\

 &   &         CE  &  2 (0.693) &2 (0.521) & 2 (0.521) &  \textbf{2 (0.394)}\\

 &  &        XB   & \textbf{2 (2.061)} & 2 (3.156) &2 (3.157) &   2 (3.176) \\

 & &         DI &2 (0.081) &   2 (0.071) &2 (0.071) &  \textbf{2 (0.082)} \\
\hline
 \multirow{4}{*}{Ovarian cancer \cite{ovar}} &  \multirow{4}{*}{2} &           PC      &                2 (0.915)                  &         2 (0.779)   & 2 (0.773)   &     \textbf{2 (0.935)} \\

&  &         CE      &       2 (0.143)         &2 (0.365)  &  2 (0.374) &       \textbf{2 (0.106) }\\

& &        XB         &     2 (2.583)            & \textbf{2 (2.072)} &  2 (2.087) &      2 (2.170)\\

& &         DI     &        2 (0.118)        & 2 (0.118) &   2 (0.118) &           \textbf{2 (0.133)}   \\
\hline
 \multirow{4}{*}{Colon cancer \cite{colon}} &  \multirow{4}{*}{2} &         PC &     2 (0.500)    &  2 (0.568) & 2 (0.568) &  \textbf{ 2 (0.733) } \\

 &  &         CE  & 2 (0.693) &  2 (0.622) & 2 (0.622) &  \textbf{2 (0.394})  \\

& &        XB   &  \textbf{2 (0.982) }&  2 (0.997) &2 (0.997)  &  2 (1.114)\\

 &  &         DI & 2 (0.311) &  \textbf{2 (0.326)} & \textbf{2 (0.326) }&   2 (0.271) \\
\hline
 \multirow{4}{*}{Ovarian cancer \cite{ova}} &  \multirow{4}{*}{2} &         PC      &  2 (0.869)        &      2 (0.779)      & 2 (0.773)  &     \textbf{2 (0.884) }\\

&  &         CE      &   2 (0.215)        & 2 (0.365) & 2 (0.374) &    \textbf{2 (0.129)}  \\

& &        XB         &    2 (1.834)       &  2 (2.072)& 2 (2.087) &    \textbf{2 (1.633)}  \\

& &         DI     &       \textbf{2 (0.165)}    &2 (0.118) &  2 (0.118)     &         2 (0.094)   \\



\hline 
 \multirow{4}{*}{Glioma \cite{data}} &  \multirow{4}{*}{4} &           PC &    4 (0.465)      &   4 (0.252)                 &       4 (0.263)       &\textbf{4 (0.637)} \\

&  &         CE&  4 (0.926)        &    4 (1.382)          &  4 (1.360) &\textbf{4 (0.664) }\\

& &        XB  &    4 (0.607)      &     4 (0.406)    &  \textbf{4 (0.405)} & 4 (0.910) \\

& &         DI &      4 (0.434)      &   \textbf{4 (0.474)}      & 4 (0.397)     &  4 (0.412)     \\
\hline
 \multirow{4}{*}{Lung cancer\cite{data}} &  \multirow{4}{*}{5} &         PC   &  5 (0.308)  &  5 (0.200)   &   5 (0.200)  & \textbf{5 (0.729)}\\

&  &         CE  &  5 (1.309)  & 5 (1.609) &   5 (1.609) &  \textbf{5 (0.497)}\\

& &        XB   &  5 (0.419) & \textbf{5 (0.292}) &     \textbf{5 (0.292)} &  5 (0.975)\\

& &         DI  &   5 (0.431) & 5 (0.342) &    5 (0.364) & \textbf{5 (0.451)} \\

\hline
 \multirow{4}{*}{Wine\cite{Dua}} &  \multirow{4}{*}{5} &         PC   &  3 (0.910)    &  3 (0.910)   &   3 (0.910)  & \textbf{3 (0.910)}\\

&  &         CE  &  3 (0.144)  & 3 (0.144) &   3 (1.33) &  \textbf{3 (0.148)}\\

& &        XB   &  \textbf{3 (6.788)} & \textbf{3 (6.788)}) &     3 (7.844) &  3 (8.431)\\

& &         DI  &    3 (0.002) & 3 (0.002) &    3 (0.004) & \textbf{3 (0.016)} \\



\hline



\end{tabular}  
\label{Table: 3}
}
\end{table*}

\begin{table}[ht]
\centering %
\caption{\textsc{  \small {Two-dimensional Gaussian synthetic dataset}}} %
\begin{tabular}{ |c|c|c|c|}
\hline 
\multicolumn{2} {|c|} {Cluster no. 1} &  \multicolumn{2} {c|} {Cluster no. 2}  \\
\hline 
$x_{1}$  &  $x_{2}$ &   $x_{1}$ &  $ x_{2}$  \\
\hline
0.53 &  $1.38$  & $6.04$ & 6.66  \\ 
\hline
1.83 &  $1.70$  & $4.80$ & 5.08 \\ 
\hline
-2.25&  $-2.76$  & $5.32$ & 5.57  \\ 
\hline
0.86 &  $2.70$  & $4.76$ & 7.05\\ 
\hline
0.31 &  $0.90$  & $ 5.22$ & 4.33 \\ 
\hline
-1.30& $-1.06$  & $5.43$ & 4.05  \\ 
\hline
 -0.43&  $-0.02$  & $4.38$ & $3.08$ \\ 
 \hline
  0.34&  $0.25$  & $5.27$ & $4.46$ \\ 
 \hline
 3.57&  $5.62$  & $5.60$ & 5.52  \\ 
\hline
2.76 &  $3.47$  & $5.09$ & 4.99 \\ 
\hline
-1.34&  $-1.25$  & $6.72$ & 7.40  \\ 
\hline
3.03 &  $3.55$  & $ 4.39$ & 4.55 \\ 
\hline
0.72 &  $0.16$  &  $4.26$ & 4.23  \\ 
\hline
-0.06& $-0.79$  & $3.25$ & $3.02$ \\ 
\hline
 0.71&  $-1.47$  & $5.91$ & $7.90$ \\ 
 \hline
  -0.20&  $0.93$  & $5.86$ & $7.35$\\ 
 \hline
  -0.12&  $0.09$  & $4.92$ & $3.76$  \\ 
 \hline
  1.48&  $1.5$  & $5.89$ & $4.33$\\ 
   \hline
  1.40&  $3.30$  & $5.18$ & $6.05$  \\ 
 \hline
  1.41&  $0.64$  & 5.29 & 3.84\\ 
\hline
\end{tabular}
\label{Table: twodataset}
\end{table}

\begin{table}[ht]
\centering %
\caption{\textsc{  \small {Results on the data set in Table \mbox {\ref{Table: twodataset}}}}} %
\begin{tabular}{ |c|c|c|c|c|    }
\hline 
& \multicolumn{2} {c|} {Cluster no. 1} &  \multicolumn{2} {c|} {Cluster no. 2}  \\
\hline 
features & $x_{1}$  &  $x_{2}$ &   $x_{1}$ &  $ x_{2}$  \\
\hline
centers &0.579   &  $0.737    $  & $4.747$ & 4.861  \\ 
\hline
weights &0.495 &  $0.505$  & $0.516$ & 0.484  \\ 
\hline
\end{tabular}
\label{Table: centertwodataset}    
\end{table}

\begin{table}[ht]
\centering %
\caption{\textsc{  \small {Four-dimensional Gaussian synthetic dataset}}} %
\begin{tabular}{ |c|c|c|c|c|c|c|c|}
\hline 
\multicolumn{4} {|c|} {Cluster no. 1} &  \multicolumn{4} {c|} {Cluster no. 2}  \\
\hline 
$x_{1}$  &  $x_{2}$ &   $x_{3}$ &  $ x_{4}$ & $x_{1}$  &  $x_{2}$ &   $x_{3}$ &  $ x_{4}$  \\
\hline
 8.15	&  6.38  &	0.54	& 1.39	& 6.05	& 4.62	& 6.67	       &0.94\\
 \hline
16.40	&9.58	&1.83	&1.71	&4.80&	4.24&	5.08&	1.44\\
\hline
14.37&	2.41&	-2.26&	-2.77&	5.33&	4.61&	5.58&	0.46\\
\hline
19.37&	6.76&	0.86&	2.71&	4.76&	7.70&	7.06&	2.88\\
\hline
10.63&	2.89&	0.32&	0.90&	5.23&	3.22&	4.33&	3.42\\
\hline
6.50&	6.72&	-1.31&	-1.07&	5.44&	7.85&	4.05&	2.73\\
\hline
2.11&	6.95&	-0.43&	-0.02&	4.38&	4.71&	3.09&	2.13\\
\hline
12.22&	0.68&	0.34&	0.25&	5.27&	0.36&	4.47&	3.22\\
\hline
15.58 &	2.55&	3.58&	5.62&	5.60&	1.76&	5.53&	3.24\\
\hline
8.47&	2.24&	2.77&	3.47&	5.09&	7.22&	4.99&	3.40\\
\hline
1.82&	6.68&	-1.35&	-1.26&	6.73&	4.73&	7.41&	3.18\\
\hline
5.33&	8.44&	3.03&	3.56&	4.39&	1.53&	4.56&	4.73\\
\hline
3.07&	3.44&	0.73&	0.16&	4.26&	3.41&	4.23&	1.04\\
\hline
5.62&	7.81&	-0.06&	-0.80&	3.25&	6.07&	3.03&	3.55\\
\hline
8.80&	6.75&	0.71&	-1.48&	5.91&	1.92&	7.91&	1.18\\
\hline
10.54&	0.07&	-0.20&	0.94&	5.87&	7.38&	7.36&	0.60\\
\hline
9.15&	6.02&	-0.12&	0.10&	4.92&	2.43&	3.77&	3.04\\
\hline
17.51&	3.87&	1.49&	1.58&	5.90&	9.17&	4.33&	2.25\\
\hline
10.36&	9.16&	1.41&	3.30&	5.18&	2.69&	6.06&	2.29\\
\hline
18.87&	0.01&	1.42&	0.64&	5.29&	7.66&	3.85&	3.31 \\
\hline
\end{tabular}
\label{Table: fourdataset}
\end{table}

\begin{table}[ht]
\centering %
\caption{\textsc{  \small {Results on the data set in Table \mbox {\ref{Table: fourdataset}}}}} %
\begin{tabular}{ |c|c|c|c|c|c|c|c|c|    }
\hline 
& \multicolumn{4} {c|} {Cluster no. 1} &  \multicolumn{4} {c|} {Cluster no. 2}  \\
\hline 
features & $x_{1}$  &  $x_{2}$ &   $x_{3}$ &  $ x_{4}$  & $x_{1}$  &  $x_{2}$ &   $x_{3}$ &  $ x_{4}$ \\
\hline
centers &6.22&	4.78	&4.40	& 2.28  & 9.37 &	4.85&	1.26&	1.05 \\ 
\hline
weights &  0.20 &  $0.19$  & $0.25$ & 0.36& 0.35& 0.03 &0.38 & 0.24   \\ 
\hline
\end{tabular}
\label{Table: centerfourdataset}    
\end{table}

\section{Results and Discussion}
\label{results and discussion}
The  proposed approach is validated on various datasets as described in Table \ref{Table: dataset} and the efficacy is presented in terms of various clustering performance measures such as accuracy rate (AR), rand index (RI) \cite{rand} and normalized mutual information (NMI) \cite{NMI}. These performance measure are mathematically expressed as follows:
\begin{align}
AR = \frac{1}{n}\sum_{j=1}^k n(k_j) 
\end{align}
\begin{align}
RI = \frac{f_1+f_2}{f_1+f_2+f_3+f_4}\\
NMI = \frac{I(X:Y)}{[H(X) + H(Y)]/2}
\end{align}
where, $n(k_{j})$ denote the number of data points correctly  assigned to cluster $j$, $n$ is the total number of data points.  A large value of  \textit{AR} represents better clustering performance. The \textit{RI} index is used to measure  similarity between the two clustering partitions. Let $k$ is the actual number of clusters, and $k^*$ is the number of clusters obtained through the various clustering methods. For a pair of points $(\text{x}_{i}, \text{x}_{j} )$, $f_1$ is the number of pairs of points that are the same in clusters $k$ and $k^{*}$, $f_2$ is the number of pairs of points that are same in  cluster $k$ and different in cluster $k^{*}$, $f_3$ is the number of pairs of points that are different in cluster $k$, and same in cluster $k^{*}$, and $f_4$ is the number of pairs of points that are different in clusters $k$, and $k^{*}.$ In the clustering,  \textit{NMI} is used to measure the information of a term to contribute for making the correct classification decision, and its values always lie  between $0$ and $1$.
    \begin{figure}
    \centering
	\includegraphics[width=8cm]{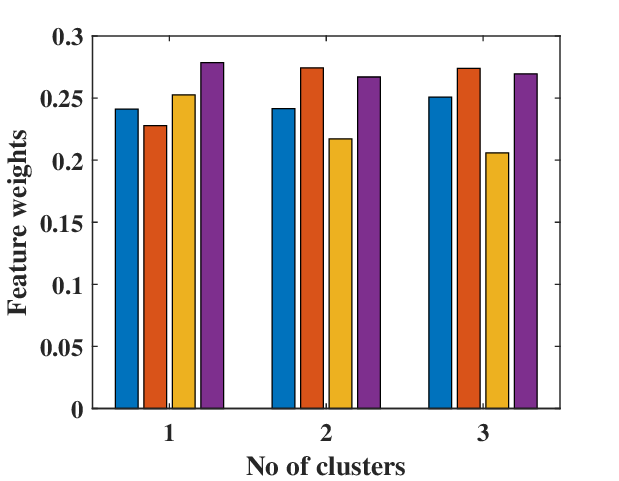}
	\caption{Feature weights for the IRIS dataset in all three clusters.}
    \label{Fig: iris weights}
    \end{figure}
The above clustering performance measures are supervised i.e., the number of cluster \textit{k } is known. However, it is generally unknown in clustering. Therefore, to find the effectiveness of proposed approach we have used different cluster validity indices, i.e., Partition Coefficient (PC) \cite{pc},  Xie and Beni's  (XB) Index \cite{xb}, Classification Entropy (CE) \cite{ce}, and Dunn's Index (DI) \cite{di}. These validity indices are mathematically written as follows:
\begin{align}
PC = \frac{1}{n}\sum_{j=1}^{k}\sum_{i=1}^{n}\mu_{ij}^2 \\
CE = -\frac{1}{n}\sum_{j=1}^{k}\sum_{i=1}^{n}\mu_{ij} \log \mu_{ij} 
\\
XB = \frac{\sum_{j=1}^{k}\sum_{i=1}^{n}\mu_{ij}^2||x_j -v_i||^2}{n\min_{ij}||x_j - v_i||^2}
\end{align}
\begin{align}
DI = \min_{i\in k}\Big\{\min{j\in k, i\neq j}\Big \{\frac{\min_{x\in k_{i}, y\in k_{j}d(x,y)}}{\max_ {i\in k}\{ \max _{x ,y \in k}d(x,y)\} }\Big\}
\end{align}

The range of PC lies in between $[1/k, 1]$,  if PC is closer to $1$ represents the best partition, whereas closer to $1/k$, the partition becomes fuzzier in the clustering. CE measures the fuzziness of the cluster partition similar to the PC, smaller the values of CE move towards the optimal cluster. In XB, the numerator represents the compactness of the fuzzy partition, and denominator denotes the strength between clusters,  smaller the values of XB move towards the optimal cluster. DI measure  the compactness in the well-separated clusters. The large value of DI represent better clustering results.

The proposed approach is validated on both small and high dimension datasets and compared with six different state-of-the-art methods, i.e.,  KM, EWKM,  AFKM, GMM, FCM, and  SCAD. The KM algorithm gives equal importance to all features in the clustering and unable to provide the optimal clusters. This problem is overcome by the EWKM clustering, where different weight are assigned to the features during the clustering. But the problem with this algorithm is in weight assignment because weight values depend on the initial cluster center. If the initial cluster center changed, the algorithm would lead to different clustering performance and unable to converge. In the AFKM, an entropy term is added in the cost function to obtain the optimal cluster in the dataset, but the problem is that it treats all features equally in the clustering. FCM assigns the soft partition of the data. Still, it gives equal importance to all features, however, in the SCAD, they consider the different features weight in the different cluster, but they do not consider the entropy which controls the partition. The objective function is simultaneously optimized in the proposed approach by taking weight entropy and fuzzy partition entropy. The weight entropy helps assign the relevant feature weight to the clustering feature, whereas the fuzzy partition helps find the optimal number of the partition. The weight control and fuzzy partition control parameter in the case of EWKM and AFKM are heuristics. However, in SCAD, weight control parameters are updated automatically in each iteration, but the fuzzy partition is not consistent. In the proposed approach, weight control and fuzzy partition control parameters are updated automatically in each iteration, and a better fuzzy partition is obtained due to relevant feature weight assignment during the clustering process.
    \begin{figure}
    \centering
	\includegraphics[width=8cm]{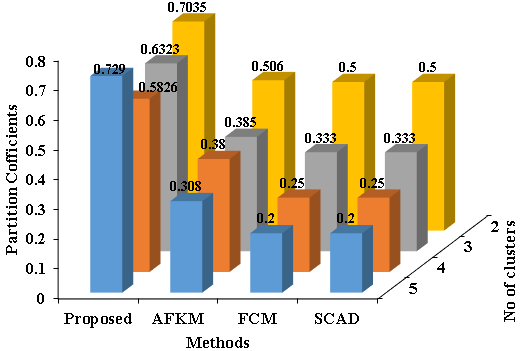}
	\caption{PCs  for the Lung dataset in varying the number of clusters.}
    \label{Fig: lung pc}
    \end{figure}

As shown in Table \ref{Table: 2}, the clustering performance of the proposed approach in terms of AR, RI, and NMI  is much better for both high dimension and non-sparse datasets in comparison with state-of-the-art methods. The cluster validity index, i.e., PC, CE, XB, and DI, are also computed and compared with state-of-the-art methods, as given in Table \ref{Table: 3}. The cluster validity measure shows that the proposed approach is able to provide the optimal clustering result with improved clustering performance. As shown in Table \ref{Table: 4}, the computational complexity of the proposed approach is slightly larger in comparison to state-of-the-art methods due to the second and third term in the objective function. However, due to these terms, the clustering performance is improved.
\begin{figure}
    \centering
	\includegraphics[width=8cm]{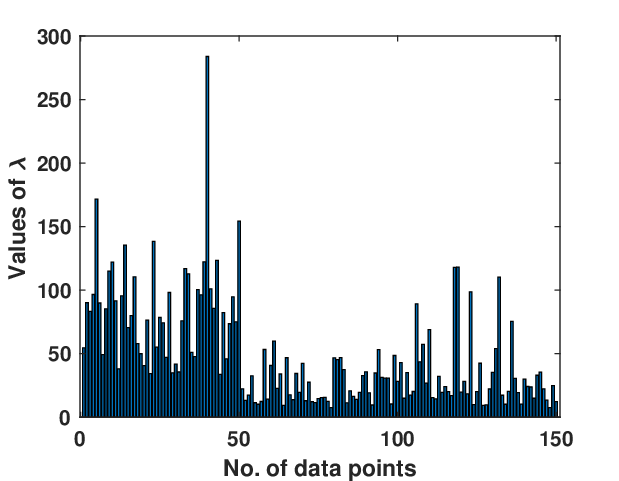}
	\caption{The values of $\lambda$ for the IRIS dataset}
\label{Fig: irislambda}
    \end{figure}
\begin{figure*}
\begin{minipage}[c]{0.95\linewidth}
 \centering
 \includegraphics[width=\linewidth]{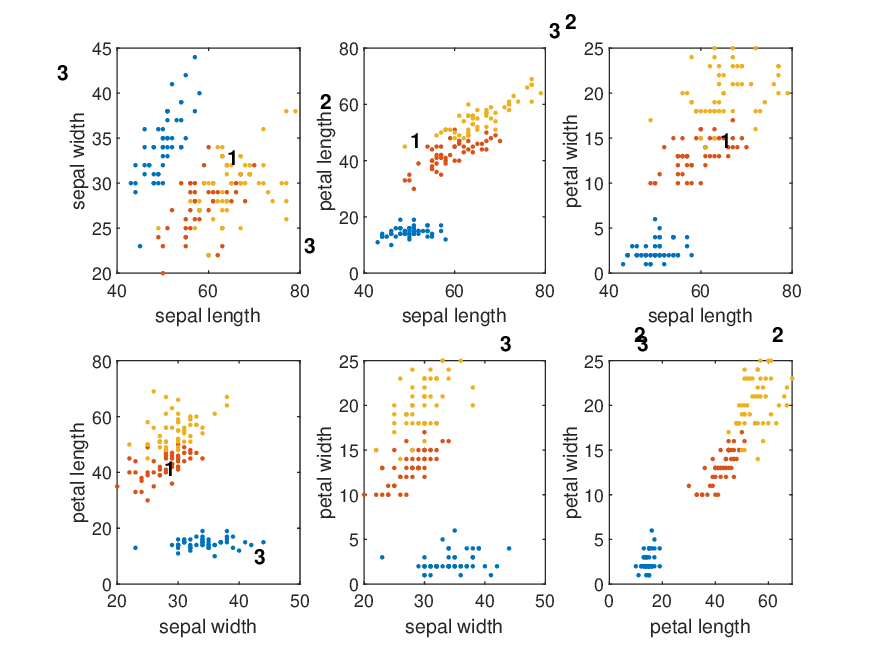}
 \caption{{Cluster centers are randomly initialized using max-min.}}
 \label{Fig: Cluster are randomly initialized}
\end{minipage}
\hfill
\begin{minipage}[c]{0.95\linewidth}
\includegraphics[width=\linewidth]{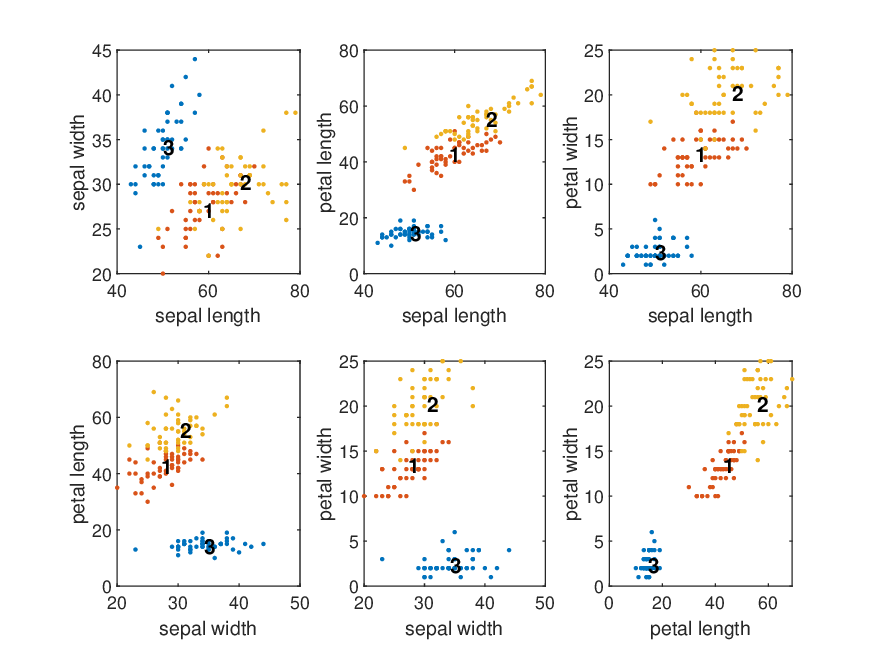}
\caption{{Cluster centers after the convergence of proposed approach. }}
\label{Fig: convergence of proposed approach }
\end{minipage}%
 \end{figure*}

\begin{figure*}
\begin{minipage}[c]{0.4\linewidth}
 \centering
 \includegraphics[width=\linewidth]{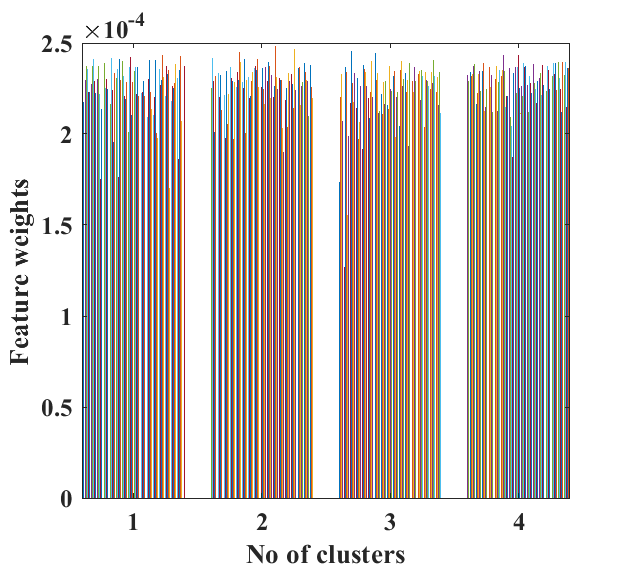}
 \caption{Feature weights for the GLIOMA dataset.}
 \label{Fig: glioma weights}
\end{minipage}
\hfill
\begin{minipage}[c]{0.6\linewidth}
\includegraphics[width=\linewidth]{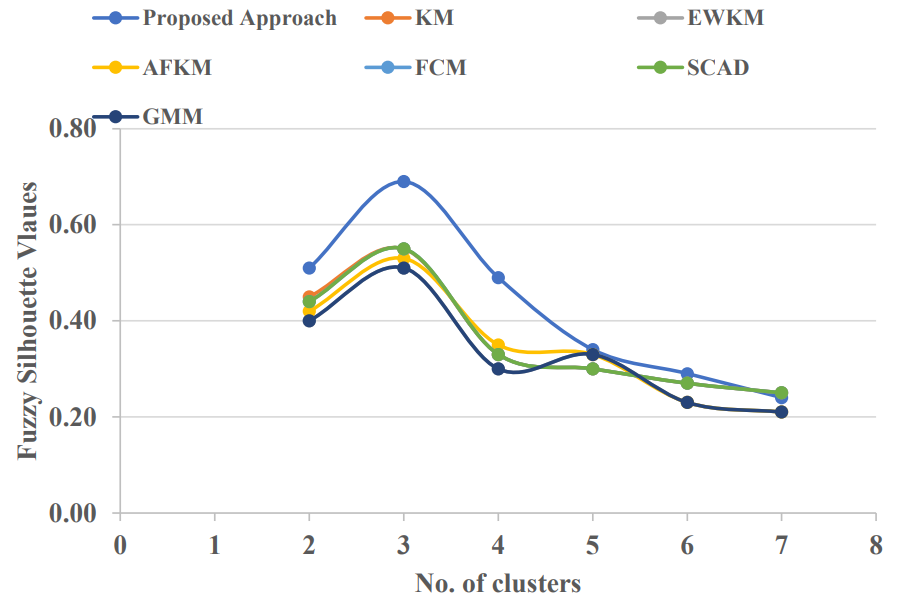}
\caption{{Fuzzy Silhouette plot for the IRIS dataset}}
\label{Fig: silhouette plot }
\end{minipage}%
 \end{figure*}
 
 The contribution of weight in each cluster corresponding to the features is also plotted for the IRIS dataset, as shown in Figures \ref{Fig: iris weights}.  As shown in Figures \ref{Fig: iris weights}, the features have different weight contribution in each cluster, which generalize the proposed approach. The fuzzy PCs are shown in Figure \ref{Fig: lung pc} by varying the number of cluster which shows that proposed approach provide the better partition.  The feature weight for the high-dimension data i.e., GLIOMA dataset is plotted to show the weight variation corresponding to the feature in the respective clusters as shown in Figure \ref{Fig: glioma weights}. For the cluster validation we have also used the  silhouette criterion and fuzzy silhouette criterion as shown in Figure \ref{Fig: silhouette plot }. The fuzzy partition control parameters are also plotted to show how parameters are changed from each iteration in Figure \ref{Fig: irislambda}. The results in Tables \ref{Table: 2} and \ref{Table: 3} are computed by running all algorithms in ten trials independently, and average results are reported. The cluster center in the proposed approach is initialized randomly using the max-min method, and initial weights of the feature are given equally important as described in Algorithm \ref{FTD-means}. The Figure \ref{Fig: Cluster are randomly initialized} represents the cluster's plot that are randomly initialized using max-min for the iris dataset. After running the algorithm twelve iterations, the cluster center's convergence is shown in Figure \ref{Fig: convergence of proposed approach }.


\begin{table}[!ht]
    \centering
   \caption{\textsc{  \small computational complexity}}\label{tab:time}
    \begin{tabular}{|c|c|}
    \hline
    Methodology  & Computational Complexity \\
    \hline
         KM \cite{km} & $O(nkm)$\\ 
         \hline
          {GMM \mbox {\cite{gmm}}} & {$O(nkm^2)$}\\ 
         \hline
         EWKM \cite{ewkm} & $O(nkm^2)$\\
         \hline
         AFKM  \cite{agglomerative}& $O(nk^2m)$  \\
         \hline
        FCM \cite{fcm}& $O(nk^2m)$\\
         \hline
         SCAD \cite{scad}  & $O(nk^2m +nkm^2)$\\
         \hline
       Proposed Approach & $O(nk^2m +nkm^2)$\\
         \hline
    \end{tabular}
    \label{Table: 4}
\end{table}

{The proposed approach effectiveness is also carried out on the synthetic datasets, as shown in Table \mbox {\ref{Table: twodataset}} and Table \mbox {\ref{Table: fourdataset}}. The Table \mbox {\ref{Table: twodataset}} contains the 20 data points of two synthetic Gaussian clusters with ($\mu_1, \sum_1)=([0, 0]^T, \textbf{I}_1)$ and ($\mu_1, \sum_1)=([5, 5]^T, \textbf{I}_2)$.  The feature weights are initialized as $\frac{1}{m}$, whereas, the cluster centers are randomly initialized using the max-min as described in Algorithm \mbox{\ref{FTD-means}}. By properly choosing the parameters $K_1$ and $K_2$, the algorithm converges after seven iterations. The obtained results are shown in Table  \mbox {\ref{Table: centertwodataset}}. Since both features are relevant, the proposed approach assigned larger weights to both features, and the estimated center are closer to those of the actual centers. To identify the relevant features, we increase the number of features to four by adding two irrelevant features to each cluster, shown in Table \mbox {\ref{Table: fourdataset}}. The first two cluster features are uniformly distributed in the intervals $[0;20]$ and $[0;10]$, respectively. Features two and four of the second cluster are uniformly distributed in the $[0;10]$ and $[0;5]$, respectively. As shown in Table \mbox {\ref{Table: centerfourdataset}}, the features are weighted during the clustering based on their importance by the algorithm after twelve iterations.}

To validate the proposed approach, an ablation study was also conducted on the IRIS \cite{Dua} dataset by assigning equal weights to features during the clustering process. The performance metrics in terms of AR, RI, and NMI are 92.00, 90.55, and 77.73; however, in terms of unsupervised performance metrics, namely PC, CE, XB, and DI, the values are 0.769, 0.390, 3.814, and 0.051. The performance of the proposed approach, as shown in Tables \ref{Table: 2} and \ref{Table: 3}, is improved by assigning appropriate weights to the features, compared to using equal weights during clustering.

\section{Conclusion}
\label{conclusion}
In this paper,  an entropy-based Variable Feature weighted fuzzy \textit{k}-means algorithm is presented for the clustering of high dimensional and sparse data with improved performance. In this approach, the objective function of the fuzzy \textit{k}-means is modified by two different entropy terms, which helps in identifying the better clustering structure for the data. The major advantage of the presented approach is that the clustering performance is consistent because it is insensitive to the initial cluster center due to different feature weights assigned to each cluster. The performance is compared with six state-of-the-art methods in terms of various clustering measures on both the real world and synthetic datasets, which shows that the proposed method is a new clustering approach to identify the number of clusters with improved performance. In the future, the correlation between the features can be considered to minimize the effect of redundant features and can be extended to categorical or mixed attributes.

\section{Compliance with Ethical Standards}
\label{sec:Compliance with Ethical Standards}

\subsection{Disclosure of Potential Conflicts of Interest}
This article does not have any funding sources and potential conflicts of interest (financial or non-financial).
\subsection{Research Involving Human Participants and/or Animals}
It also does not have any research or data related to human participants considering animal welfare.
\subsection{Informed Consent}
The authors have given their consent to submit this article in this prestigious journal.
\subsection{Data Availability}
The datasets analysed during the current study are available at repositories \url{http://featureselection.asu.edu/datasets.php}, \url{https://in.mathworks.com/help/stats/sample-data-sets.html}, \url{http://cilab.ujn.edu.cn/datasets.htm} and \url{http://archive.ics.uci.edu/ml}.

\bibliography{References}


\begin{thebibliography}{44}
\ifx \bisbn   \undefined \def \bisbn  #1{ISBN #1}\fi
\ifx \binits  \undefined \def \binits#1{#1}\fi
\ifx \bauthor  \undefined \def \bauthor#1{#1}\fi
\ifx \batitle  \undefined \def \batitle#1{#1}\fi
\ifx \bjtitle  \undefined \def \bjtitle#1{#1}\fi
\ifx \bvolume  \undefined \def \bvolume#1{\textbf{#1}}\fi
\ifx \byear  \undefined \def \byear#1{#1}\fi
\ifx \bissue  \undefined \def \bissue#1{#1}\fi
\ifx \bfpage  \undefined \def \bfpage#1{#1}\fi
\ifx \blpage  \undefined \def \blpage #1{#1}\fi
\ifx \burl  \undefined \def \burl#1{\textsf{#1}}\fi
\ifx \doiurl  \undefined \def \doiurl#1{\url{https://doi.org/#1}}\fi
\ifx \betal  \undefined \def \betal{\textit{et al.}}\fi
\ifx \binstitute  \undefined \def \binstitute#1{#1}\fi
\ifx \binstitutionaled  \undefined \def \binstitutionaled#1{#1}\fi
\ifx \bctitle  \undefined \def \bctitle#1{#1}\fi
\ifx \beditor  \undefined \def \beditor#1{#1}\fi
\ifx \bpublisher  \undefined \def \bpublisher#1{#1}\fi
\ifx \bbtitle  \undefined \def \bbtitle#1{#1}\fi
\ifx \bedition  \undefined \def \bedition#1{#1}\fi
\ifx \bseriesno  \undefined \def \bseriesno#1{#1}\fi
\ifx \blocation  \undefined \def \blocation#1{#1}\fi
\ifx \bsertitle  \undefined \def \bsertitle#1{#1}\fi
\ifx \bsnm \undefined \def \bsnm#1{#1}\fi
\ifx \bsuffix \undefined \def \bsuffix#1{#1}\fi
\ifx \bparticle \undefined \def \bparticle#1{#1}\fi
\ifx \barticle \undefined \def \barticle#1{#1}\fi
\bibcommenthead
\ifx \bconfdate \undefined \def \bconfdate #1{#1}\fi
\ifx \botherref \undefined \def \botherref #1{#1}\fi
\ifx \url \undefined \def \url#1{\textsf{#1}}\fi
\ifx \bchapter \undefined \def \bchapter#1{#1}\fi
\ifx \bbook \undefined \def \bbook#1{#1}\fi
\ifx \bcomment \undefined \def \bcomment#1{#1}\fi
\ifx \oauthor \undefined \def \oauthor#1{#1}\fi
\ifx \citeauthoryear \undefined \def \citeauthoryear#1{#1}\fi
\ifx \endbibitem  \undefined \def \endbibitem {}\fi
\ifx \bconflocation  \undefined \def \bconflocation#1{#1}\fi
\ifx \arxivurl  \undefined \def \arxivurl#1{\textsf{#1}}\fi
\csname PreBibitemsHook\endcsname

\bibitem[\protect\citeauthoryear{Hennig et~al.}{2015}]{HCA}
\begin{bbook}
\bauthor{\bsnm{Hennig}, \binits{C.}},
\bauthor{\bsnm{Meila}, \binits{M.}},
\bauthor{\bsnm{Murtagh}, \binits{F.}},
\bauthor{\bsnm{Rocci}, \binits{R.}}:
\bbtitle{Handbook of Cluster Analysis}.
\bpublisher{CRC Press}, \blocation{???}
(\byear{2015})
\end{bbook}
\endbibitem

\bibitem[\protect\citeauthoryear{Verma et~al.}{2021}]{bideal}
\begin{barticle}
\bauthor{\bsnm{Verma}, \binits{N.K.}},
\bauthor{\bsnm{Sharma}, \binits{T.}},
\bauthor{\bsnm{Dixit}, \binits{S.}},
\bauthor{\bsnm{Agrawal}, \binits{P.}},
\bauthor{\bsnm{Sengupta}, \binits{S.}},
\bauthor{\bsnm{Singh}, \binits{V.}}:
\batitle{Bideal: a toolbox for bicluster analysis—generation, visualization
  and validation}.
\bjtitle{SN Computer Science}
\bvolume{2},
\bfpage{1}--\blpage{15}
(\byear{2021})
\end{barticle}
\endbibitem

\bibitem[\protect\citeauthoryear{Bezdek}{1981}]{bezdek}
\begin{bbook}
\bauthor{\bsnm{Bezdek}, \binits{J.C.}}:
\bbtitle{Pattern Recognition with Fuzzy Objective Function Algorithms}.
\bpublisher{New York: Plenum}, \blocation{???}
(\byear{1981})
\end{bbook}
\endbibitem

\bibitem[\protect\citeauthoryear{Karypis et~al.}{1999}]{hierarchical}
\begin{botherref}
\oauthor{\bsnm{Karypis}, \binits{G.}},
\oauthor{\bsnm{Han}, \binits{E.-H.S.}},
\oauthor{\bsnm{Kumar}, \binits{V.}}:
Chameleon: Hierarchical clustering using dynamic modeling.
Computer
(8),
68--75
(1999)
\end{botherref}
\endbibitem

\bibitem[\protect\citeauthoryear{Kriegel et~al.}{2011}]{density}
\begin{barticle}
\bauthor{\bsnm{Kriegel}, \binits{H.-P.}},
\bauthor{\bsnm{Kr{\"o}ger}, \binits{P.}},
\bauthor{\bsnm{Sander}, \binits{J.}},
\bauthor{\bsnm{Zimek}, \binits{A.}}:
\batitle{Density-based clustering}.
\bjtitle{Wiley Interdisciplinary Reviews: Data Mining and Knowledge Discovery}
\bvolume{1}(\bissue{3}),
\bfpage{231}--\blpage{240}
(\byear{2011})
\end{barticle}
\endbibitem

\bibitem[\protect\citeauthoryear{Abhadiomhen et~al.}{2024}]{spectral}
\begin{barticle}
\bauthor{\bsnm{Abhadiomhen}, \binits{S.E.}},
\bauthor{\bsnm{Ezeora}, \binits{N.J.}},
\bauthor{\bsnm{Ganaa}, \binits{E.D.}},
\bauthor{\bsnm{Nzeh}, \binits{R.C.}},
\bauthor{\bsnm{Adeyemo}, \binits{I.}},
\bauthor{\bsnm{Uzo}, \binits{I.U.}},
\bauthor{\bsnm{Oguike}, \binits{O.}}:
\batitle{Spectral type subspace clustering methods: multi-perspective
  analysis}.
\bjtitle{Multimedia Tools and Applications}
\bvolume{83}(\bissue{16}),
\bfpage{47455}--\blpage{47475}
(\byear{2024})
\end{barticle}
\endbibitem

\bibitem[\protect\citeauthoryear{Ordonez and Omiecinski}{2002}]{grid}
\begin{bchapter}
\bauthor{\bsnm{Ordonez}, \binits{C.}},
\bauthor{\bsnm{Omiecinski}, \binits{E.}}:
\bctitle{{FREM}: fast and robust {EM} clustering for large data sets}.
In: \bbtitle{Proceedings of the Eleventh International Conference on
  Information and Knowledge Mgmt.},
pp. \bfpage{590}--\blpage{599}
(\byear{2002}).
\bcomment{ACM}
\end{bchapter}
\endbibitem

\bibitem[\protect\citeauthoryear{Jain and C.}{1988}]{Ac}
\begin{bbook}
\bauthor{\bsnm{Jain}, \binits{A.K.}},
\bauthor{\bsnm{C.}, \binits{D.R.}}:
\bbtitle{Algorithms for Clustering Data}.
\bpublisher{Prentice Hall}, \blocation{???}
(\byear{1988})
\end{bbook}
\endbibitem

\bibitem[\protect\citeauthoryear{Anderberg}{1988}]{Ad}
\begin{bbook}
\bauthor{\bsnm{Anderberg}, \binits{M.R.}}:
\bbtitle{Clustering Analysis for Applications}.
\bpublisher{Academic}, \blocation{???}
(\byear{1988})
\end{bbook}
\endbibitem

\bibitem[\protect\citeauthoryear{Ball and Hall}{1967}]{ball}
\begin{barticle}
\bauthor{\bsnm{Ball}, \binits{G.H.}},
\bauthor{\bsnm{Hall}, \binits{D.J.}}:
\batitle{A clustering technique for summarizing multivariate data}.
\bjtitle{Behavioral science}
\bvolume{12}(\bissue{2}),
\bfpage{153}--\blpage{155}
(\byear{1967})
\end{barticle}
\endbibitem

\bibitem[\protect\citeauthoryear{Ruspini}{1969}]{ruspini}
\begin{barticle}
\bauthor{\bsnm{Ruspini}, \binits{E.H.}}:
\batitle{A new approach to clustering}.
\bjtitle{Information and Control}
\bvolume{15}(\bissue{1}),
\bfpage{22}--\blpage{32}
(\byear{1969})
\end{barticle}
\endbibitem

\bibitem[\protect\citeauthoryear{Bezdek}{1980}]{bezdek1980}
\begin{botherref}
\oauthor{\bsnm{Bezdek}, \binits{J.C.}}:
A convergence theorem for the fuzzy isodata clustering algorithms.
IEEE Transactions on Pattern Analysis \& Machine Intelligence
(1),
1--8
(1980)
\end{botherref}
\endbibitem

\bibitem[\protect\citeauthoryear{Singh et~al.}{2019}]{viknano}
\begin{barticle}
\bauthor{\bsnm{Singh}, \binits{V.}},
\bauthor{\bsnm{Verma}, \binits{N.K.}},
\bauthor{\bsnm{Cui}, \binits{Y.}}:
\batitle{Type-2 fuzzy pca approach in extracting salient features for molecular
  cancer diagnostics and prognostics}.
\bjtitle{IEEE Transactions on Nanobioscience}
\bvolume{18}(\bissue{3}),
\bfpage{482}--\blpage{489}
(\byear{2019})
\end{barticle}
\endbibitem

\bibitem[\protect\citeauthoryear{Singh et~al.}{2024}]{ieetfs}
\begin{botherref}
\oauthor{\bsnm{Singh}, \binits{V.}},
\oauthor{\bsnm{Pal}, \binits{V.C.}},
\oauthor{\bsnm{Pati}, \binits{A.}}, et al.:
Adaptive type-2 fuzzy filter with kernel density estimation for impulse noise
  removal.
IEEE Transactions on Fuzzy Systems
(2024)
\end{botherref}
\endbibitem

\bibitem[\protect\citeauthoryear{Bezdek et~al.}{1984}]{fcm}
\begin{barticle}
\bauthor{\bsnm{Bezdek}, \binits{J.C.}},
\bauthor{\bsnm{Ehrlich}, \binits{R.}},
\bauthor{\bsnm{Full}, \binits{W.}}:
\batitle{{FCM}: The fuzzy c-means clustering algorithm}.
\bjtitle{Computers \& Geosciences}
\bvolume{10}(\bissue{2-3}),
\bfpage{191}--\blpage{203}
(\byear{1984})
\end{barticle}
\endbibitem

\bibitem[\protect\citeauthoryear{Zhu et~al.}{2009}]{Imfcm}
\begin{barticle}
\bauthor{\bsnm{Zhu}, \binits{L.}},
\bauthor{\bsnm{Chung}, \binits{F.L.}},
\bauthor{\bsnm{Wang}, \binits{S.}}:
\batitle{Generalized fuzzy c-means clustering algorithm with improved fuzzy
  partitions}.
\bjtitle{IEEE Transactions on Systems, Man, and Cybernetics, Part B
  (Cybernetics)}
\bvolume{39}(\bissue{3}),
\bfpage{578}--\blpage{591}
(\byear{2009})
\end{barticle}
\endbibitem

\bibitem[\protect\citeauthoryear{Huang et~al.}{2005}]{wkm}
\begin{botherref}
\oauthor{\bsnm{Huang}, \binits{J.Z.}},
\oauthor{\bsnm{Ng}, \binits{M.K.}},
\oauthor{\bsnm{Rong}, \binits{H.}},
\oauthor{\bsnm{Li}, \binits{Z.}}:
Automated variable weighting in k-means type clustering.
IEEE Transactions on Pattern Analysis \& Machine Intelligence
(5),
657--668
(2005)
\end{botherref}
\endbibitem

\bibitem[\protect\citeauthoryear{Jing et~al.}{2007}]{ewkm}
\begin{botherref}
\oauthor{\bsnm{Jing}, \binits{L.}},
\oauthor{\bsnm{Ng}, \binits{M.K.}},
\oauthor{\bsnm{Huang}, \binits{J.Z.}}:
An entropy weighting k-means algorithm for subspace clustering of
  high-dimensional sparse data.
IEEE Transactions on Knowledge \& Data Engineering
(8),
1026--1041
(2007)
\end{botherref}
\endbibitem

\bibitem[\protect\citeauthoryear{Witten and Tibshirani}{2010}]{fs}
\begin{barticle}
\bauthor{\bsnm{Witten}, \binits{D.M.}},
\bauthor{\bsnm{Tibshirani}, \binits{R.}}:
\batitle{A framework for feature selection in clustering}.
\bjtitle{Journal of the American Statistical Association}
\bvolume{105}(\bissue{490}),
\bfpage{713}--\blpage{726}
(\byear{2010})
\end{barticle}
\endbibitem

\bibitem[\protect\citeauthoryear{Wang et~al.}{2004}]{wfcm}
\begin{barticle}
\bauthor{\bsnm{Wang}, \binits{X.}},
\bauthor{\bsnm{Wang}, \binits{Y.}},
\bauthor{\bsnm{Wang}, \binits{L.}}:
\batitle{Improving fuzzy c-means clustering based on feature-weight learning}.
\bjtitle{Pattern Recognition Letters}
\bvolume{25}(\bissue{10}),
\bfpage{1123}--\blpage{1132}
(\byear{2004})
\end{barticle}
\endbibitem

\bibitem[\protect\citeauthoryear{Frigui and Nasraoui}{2004}]{scad}
\begin{barticle}
\bauthor{\bsnm{Frigui}, \binits{H.}},
\bauthor{\bsnm{Nasraoui}, \binits{O.}}:
\batitle{Unsupervised learning of prototypes and attribute weights}.
\bjtitle{Pattern Recognition}
\bvolume{37}(\bissue{3}),
\bfpage{567}--\blpage{581}
(\byear{2004})
\end{barticle}
\endbibitem

\bibitem[\protect\citeauthoryear{Yang and Nataliani}{2017}]{frfcm}
\begin{barticle}
\bauthor{\bsnm{Yang}, \binits{S.} \bsuffix{Miin}},
\bauthor{\bsnm{Nataliani}, \binits{Y.}}:
\batitle{A feature-reduction fuzzy clustering algorithm based on
  feature-weighted entropy}.
\bjtitle{IEEE Transactions on Fuzzy Systems}
\bvolume{26}(\bissue{2}),
\bfpage{817}--\blpage{835}
(\byear{2017})
\end{barticle}
\endbibitem

\bibitem[\protect\citeauthoryear{Singh and Verma}{2022}]{singh}
\begin{barticle}
\bauthor{\bsnm{Singh}, \binits{V.}},
\bauthor{\bsnm{Verma}, \binits{N.K.}}:
\batitle{Gene expression data analysis using feature weighted robust
  fuzzy-means clustering}.
\bjtitle{IEEE Transactions on NanoBioscience}
\bvolume{22}(\bissue{1}),
\bfpage{99}--\blpage{105}
(\byear{2022})
\end{barticle}
\endbibitem

\bibitem[\protect\citeauthoryear{Modha and Spangler}{2003}]{fwk}
\begin{barticle}
\bauthor{\bsnm{Modha}, \binits{D.S.}},
\bauthor{\bsnm{Spangler}, \binits{W.S.}}:
\batitle{Feature weighting in k-means clustering}.
\bjtitle{Machine Learning}
\bvolume{52}(\bissue{3}),
\bfpage{217}--\blpage{237}
(\byear{2003})
\end{barticle}
\endbibitem

\bibitem[\protect\citeauthoryear{Singh et~al.}{2024}]{nor}
\begin{botherref}
\oauthor{\bsnm{Singh}, \binits{V.}},
\oauthor{\bsnm{Kirtipal}, \binits{N.}},
\oauthor{\bsnm{Lim}, \binits{S.}},
\oauthor{\bsnm{Lee}, \binits{S.}}:
Normalization of single-cell rna-seq data using partial least squares with
  adaptive fuzzy weight.
bioRxiv,
2024--08
(2024)
\end{botherref}
\endbibitem

\bibitem[\protect\citeauthoryear{Keller and Klawonn}{2000}]{wfc}
\begin{barticle}
\bauthor{\bsnm{Keller}, \binits{A.}},
\bauthor{\bsnm{Klawonn}, \binits{F.}}:
\batitle{Fuzzy clustering with weighting of data variables}.
\bjtitle{International Journal of Uncertainty, Fuzziness and Knowledge-Based
  Systems}
\bvolume{8}(\bissue{06}),
\bfpage{735}--\blpage{746}
(\byear{2000})
\end{barticle}
\endbibitem

\bibitem[\protect\citeauthoryear{Borgelt}{2009}]{fsc}
\begin{bchapter}
\bauthor{\bsnm{Borgelt}, \binits{C.}}:
\bctitle{Fuzzy subspace clustering}.
In: \bbtitle{Advances in Data Analysis, Data Handling and Business
  Intelligence},
pp. \bfpage{93}--\blpage{103}.
\bpublisher{Springer}, \blocation{???}
(\byear{2009})
\end{bchapter}
\endbibitem

\bibitem[\protect\citeauthoryear{Sahbi and Boujemaa}{2005}]{fcer}
\begin{bchapter}
\bauthor{\bsnm{Sahbi}, \binits{H.}},
\bauthor{\bsnm{Boujemaa}, \binits{N.}}:
\bctitle{Fuzzy clustering: Consistency of entropy regularization}.
In: \bbtitle{Computational Intelligence, Theory and Applications},
pp. \bfpage{95}--\blpage{107}.
\bpublisher{Springer}, \blocation{???}
(\byear{2005})
\end{bchapter}
\endbibitem

\bibitem[\protect\citeauthoryear{Guillon et~al.}{2019}]{pfsc}
\begin{barticle}
\bauthor{\bsnm{Guillon}, \binits{A.}},
\bauthor{\bsnm{Lesot}, \binits{M.-J.}},
\bauthor{\bsnm{Marsala}, \binits{C.}}:
\batitle{A proximal framework for fuzzy subspace clustering}.
\bjtitle{Fuzzy Sets and Systems}
\bvolume{366},
\bfpage{34}--\blpage{45}
(\byear{2019})
\end{barticle}
\endbibitem

\bibitem[\protect\citeauthoryear{Chen}{2011}]{tw}
\begin{barticle}
\bauthor{\bsnm{Chen}, \binits{X.} \bsuffix{\textit{et al.}}}:
\batitle{{TW}-k-means: Automated two-level variable weighting clustering
  algorithm for multiview data}.
\bjtitle{IEEE Transactions on Knowledge \& Data Engineering}
\bvolume{25}(\bissue{4}),
\bfpage{932}--\blpage{944}
(\byear{2011})
\end{barticle}
\endbibitem

\bibitem[\protect\citeauthoryear{Li et~al.}{2008}]{agglomerative}
\begin{barticle}
\bauthor{\bsnm{Li}, \binits{M.J.}}, \betal:
\batitle{Agglomerative fuzzy k-means clustering algorithm with selection of
  number of clusters}.
\bjtitle{IEEE Transactions on Knowledge \& Data Engineering}
\bvolume{20}(\bissue{11}),
\bfpage{1519}--\blpage{1534}
(\byear{2008})
\end{barticle}
\endbibitem

\bibitem[\protect\citeauthoryear{Dua and Graff}{2017}]{Dua}
\begin{botherref}
\oauthor{\bsnm{Dua}, \binits{D.}},
\oauthor{\bsnm{Graff}, \binits{C.}}:
{UCI} Machine Learning Repository
(2017).
\url{http://archive.ics.uci.edu/ml}
\end{botherref}
\endbibitem

\bibitem[\protect\citeauthoryear{}{}]{ovar}
\begin{botherref}
\url{http://cilab.ujn.edu.cn/datasets.htm}.
[Online; accessed 01-Nov-2019]
\end{botherref}
\endbibitem

\bibitem[\protect\citeauthoryear{Alon}{1999}]{colon}
\begin{barticle}
\bauthor{\bsnm{Alon}, \binits{U.} \bsuffix{\textit{et al.}}}:
\batitle{Broad patterns of gene expression revealed by clustering analysis of
  tumor and normal colon tissues probed by oligonucleotide arrays}.
\bjtitle{Proceedings of the National Academy of Sciences}
\bvolume{96}(\bissue{12}),
\bfpage{6745}--\blpage{6750}
(\byear{1999})
\end{barticle}
\endbibitem

\bibitem[\protect\citeauthoryear{}{}]{ova}
\begin{botherref}
\url{https://in.mathworks.com/help/stats/sample-data-sets.html}
\end{botherref}
\endbibitem

\bibitem[\protect\citeauthoryear{}{}]{data}
\begin{botherref}
\url{http://featureselection.asu.edu/datasets.php}
\end{botherref}
\endbibitem

\bibitem[\protect\citeauthoryear{MacQueen et~al.}{1967}]{km}
\begin{bchapter}
\bauthor{\bsnm{MacQueen}, \binits{J.}}, \betal:
\bctitle{Some methods for classification and analysis of multivariate
  observations}.
In: \bbtitle{Proceedings of the Fifth Berkeley Symposium on Mathematical
  Statistics and Probability},
vol. \bseriesno{1},
pp. \bfpage{281}--\blpage{297}
(\byear{1967}).
\bcomment{Oakland, CA, USA}
\end{bchapter}
\endbibitem

\bibitem[\protect\citeauthoryear{Verbeek et~al.}{2003}]{gmm}
\begin{barticle}
\bauthor{\bsnm{Verbeek}, \binits{J.J.}},
\bauthor{\bsnm{Vlassis}, \binits{N.}},
\bauthor{\bsnm{Kr{\"o}se}, \binits{B.}}:
\batitle{Efficient greedy learning of gaussian mixture models}.
\bjtitle{Neural computation}
\bvolume{15}(\bissue{2}),
\bfpage{469}--\blpage{485}
(\byear{2003})
\end{barticle}
\endbibitem

\bibitem[\protect\citeauthoryear{Rand}{1971}]{rand}
\begin{barticle}
\bauthor{\bsnm{Rand}, \binits{W.M.}}:
\batitle{Objective criteria for the evaluation of clustering methods}.
\bjtitle{Journal of the American Statistical Association}
\bvolume{66}(\bissue{336}),
\bfpage{846}--\blpage{850}
(\byear{1971})
\end{barticle}
\endbibitem

\bibitem[\protect\citeauthoryear{He et~al.}{2006}]{NMI}
\begin{bchapter}
\bauthor{\bsnm{He}, \binits{X.}},
\bauthor{\bsnm{Cai}, \binits{D.}},
\bauthor{\bsnm{Niyogi}, \binits{P.}}:
\bctitle{Laplacian score for feature selection}.
In: \bbtitle{Advances in Neural Information Processing Systems},
pp. \bfpage{507}--\blpage{514}
(\byear{2006})
\end{bchapter}
\endbibitem

\bibitem[\protect\citeauthoryear{Bezdek}{1974}]{pc}
\begin{barticle}
\bauthor{\bsnm{Bezdek}, \binits{J.C.}}:
\batitle{Numerical taxonomy with fuzzy sets}.
\bjtitle{Journal of Mathematical Biology}
\bvolume{1}(\bissue{1}),
\bfpage{57}--\blpage{71}
(\byear{1974})
\end{barticle}
\endbibitem

\bibitem[\protect\citeauthoryear{Xie and Beni}{1991}]{xb}
\begin{botherref}
\oauthor{\bsnm{Xie}, \binits{X.L.}},
\oauthor{\bsnm{Beni}, \binits{G.}}:
A validity measure for fuzzy clustering.
IEEE Transactions on Pattern Analysis \& Machine Intelligence
(8),
841--847
(1991)
\end{botherref}
\endbibitem

\bibitem[\protect\citeauthoryear{Bezdek}{1973}]{ce}
\begin{botherref}
\oauthor{\bsnm{Bezdek}, \binits{J.C.}}:
Cluster validity with fuzzy sets
(1973)
\end{botherref}
\endbibitem

\bibitem[\protect\citeauthoryear{Dunn}{1974}]{di}
\begin{barticle}
\bauthor{\bsnm{Dunn}, \binits{J.C.}}:
\batitle{Well-separated clusters and optimal fuzzy partitions}.
\bjtitle{Journal of Cybernetics}
\bvolume{4}(\bissue{1}),
\bfpage{95}--\blpage{104}
(\byear{1974})
\end{barticle}
\endbibitem

\end{thebibliography}

\end{document}